%% file: arxiv.tex
\documentclass{article}
\usepackage{ijcai23}

\usepackage{times}
\usepackage{soul,xcolor,colortbl}
\usepackage{url}
\definecolor{cvprblue}{rgb}{0.21,0.49,0.74}
\definecolor{LightCyan}{rgb}{0.88,1,1}
\usepackage[colorlinks=False,citecolor=cvprblue,linkcolor=red]{hyperref}
\usepackage[utf8]{inputenc}
\usepackage[small]{caption}
\usepackage{graphicx}
\usepackage{multirow}
\usepackage{tabularx}
\usepackage{booktabs}
\usepackage{algorithm}
\usepackage{algorithmic}
\usepackage[switch]{lineno}
\usepackage{amsmath,amssymb,bm,bbm} 
\usepackage{todonotes}
\usepackage{makecell}
\usepackage{pict2e} 

\newcommand{\myPara}[1]{\vspace{.05in}\noindent\textbf{#1}}

\def\ie{\textit{i.e.}}

\usepackage{enumitem}
\usepackage{overpic}
\usepackage{makecell}

\usepackage[capitalize,noabbrev]{cleveref}
\crefname{section}{Sect.}{Sects.}
\crefname{figure}{Fig.}{Figs.}
\crefname{table}{Tab.}{Tabs.}
\crefname{equation}{Eq.}{Eqs.}
\crefname{algorithm}{Alg.}{Algs.}

\newtheorem{definition}{Definition}
\usepackage[misc]{ifsym} 

\title{
CoFInAl: Enhancing Action Quality Assessment \\ with Coarse-to-Fine Instruction Alignment
}

\author{
Kanglei Zhou$^{\text{1}}$\and
Junlin Li$^{\text{2}}$\and
Ruizhi Cai$^{\text{1}}$\and  \\
Liyuan Wang$^{\text{3}}$\and 
Xingxing Zhang$^{\text{3}}$\And
Xiaohui Liang$^{{\text{1,4}}~\textrm{\Letter}}$
\affiliations 
$^{\text{1}}$ State Key Laboratory of Virtual Reality Technology and Systems, Beihang University\\
$^{\text{2}}$ China Three Gorges University \quad $^{\text{4}}$ Zhongguancun Laboratory \\
$^{\text{3}}$ Department of Computer Science and Technology, Institute for AI, BNRist Center, Tsinghua-Bosch Joint ML Center, THBI Lab, Tsinghua University\\
\emails
{\tt liang\_xiaohui@buaa.edu.cn}
}

\begin{document}

\maketitle

\begin{abstract}
    Action Quality Assessment (AQA) is pivotal for quantifying actions across domains like sports and medical care. Existing methods often rely on pre-trained backbones from large-scale action recognition datasets to boost performance on smaller AQA datasets.
    However, this common strategy yields suboptimal results due to the inherent struggle of these backbones to capture the subtle cues essential for AQA. Moreover, fine-tuning on smaller datasets risks overfitting.
    To address these issues, we propose Coarse-to-Fine Instruction Alignment (CoFInAl).
    Inspired by recent advances in large language model tuning, CoFInAl aligns AQA with broader pre-trained tasks by reformulating it as a coarse-to-fine classification task. 
    Initially, it learns grade prototypes for coarse assessment and then utilizes fixed sub-grade prototypes for fine-grained assessment. This hierarchical approach mirrors the judging process, enhancing interpretability within the AQA framework. 
    Experimental results on two long-term AQA datasets demonstrate CoFInAl achieves state-of-the-art performance with significant correlation gains of 5.49\% and 3.55\% on Rhythmic Gymnastics and Fis-V, respectively. 
    \footnote{Code is available at \href{https://github.com/ZhouKanglei/CoFInAl_AQA}{github.com/ZhouKanglei/CoFInAl\_AQA}.}

\end{abstract}

\section{Introduction}
Action Quality Assessment (AQA) aims to evaluate the quality of executed actions and is commonly used for analyzing human movements and activities \cite{wang2021survey}. Its application spans diverse domains, including sports \cite{liu2023figure,zhou2023hierarchical,zhou2024magr}, medical care \cite{zhou2023video}, surgical training \cite{ding2023sedskill}, \emph{etc}.

\begin{figure}
    \centering
    \includegraphics[width=\linewidth,trim=80 160 80 155,clip]{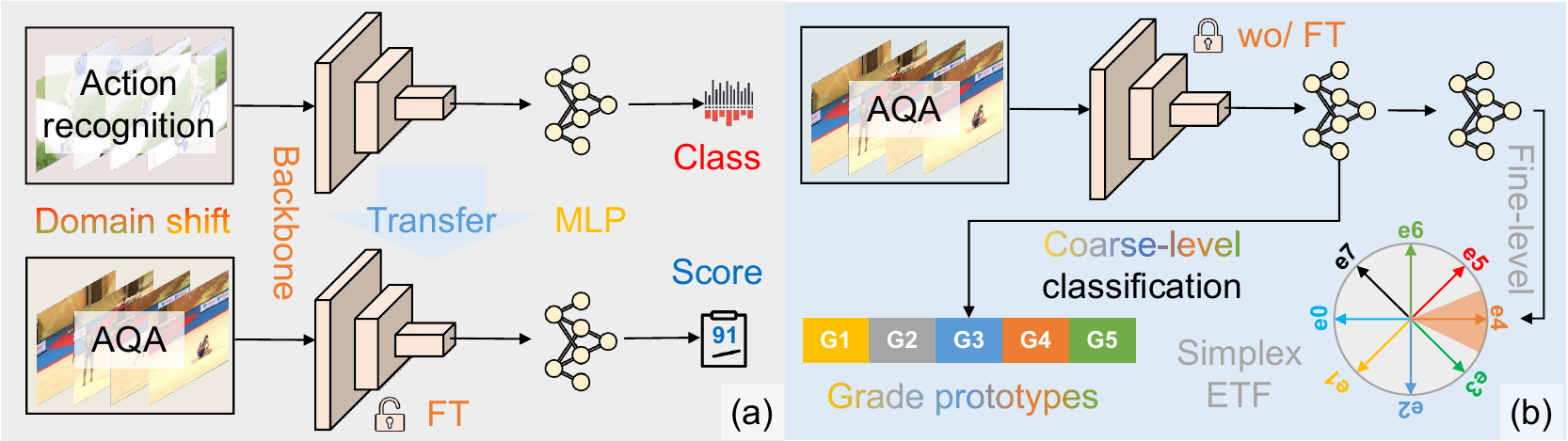}
    \caption{
    Motivation: (a) Previous methods often fine-tune large-scale pre-trained action recognition backbones, yielding suboptimal performance due to domain shift and overfitting. (b) Our method aligns AQA with broader tasks via coarse-to-fine instruction alignment, employing grade prototype learning and fine-grained sub-grade classification with a simplex Equiangular Tight Frame (ETF).
    }
    \label{fig:motivation}
\end{figure}

The primary challenge in AQA originates in the scarcity of labeled data, resulting in the typically small size of AQA datasets. For instance, the representative MTL-AQA dataset \cite{parmar2019and} comprises only about 1,000 samples. To improve the performance of AQA, a prevalent strategy \cite{parmar2019action,yu2021group,bai2022action} involves leveraging the backbone pre-trained on large-scale action recognition datasets (\emph{e.g.}, Kinetics 400 \cite{kay2017kinetics} with over 300,000 samples) to adapt the score regression requirements of small-scale AQA datasets (see \cref{fig:motivation}(a)). However, this strategy faces two critical issues. First, the \textbf{domain shift} between the pre-trained classification task of action recognition and the fine-tuned regression task of AQA renders the pre-trained model suboptimal for AQA. Second, fine-tuning the pre-trained model on small-scale AQA datasets often involves a severe risk of \textbf{overfitting}, making it difficult to bridge the domain shift. 

To overcome the double challenges of domain shift and overfitting that constrain the AQA performance, we propose here an innovative approach named Coarse-to-Fine Instruction Alignment (CoFInAl), which aligns the objectives of pre-training and fine-tuning through characterizing AQA as a coarse-to-fine classification task (see \cref{fig:motivation}(b)).
This coarse-to-fine process mirrors the two-step assessment taken by a judge, initially identifying a coarse grade and subsequently discerning variations within each grade.

At the coarse level, we categorize the performance of actions into different grades, such as excellent, good, or fair, representing varying levels of skill. For instance, in figure skating, a routine may include a flawless triple axel (excellent), graceful spins (good), and an awkward landing (fair). To identify the grade of a given sample, we propose the Grade Parsing Module (GPM, \cref{sec:gpm}), which dynamically captures performance criteria for the same action type through a set of grade prototypes. These learned criteria are then utilized to query each action step, resulting in score responses as coarse-grained features. Subsequently, these features are categorized using an MLP classifier to predict the coarse-grained grades. However, relying solely on coarse-level classification is insufficient for effectively distinguishing intra-grade actions, as neural collapse \cite{papyan2020prevalence} suggests that last-layer features of the same class collapse into their intra-class mean. This motivates us to introduce a fine-grained classification task to discern variations within each grade.

At the fine level, we categorize the performance of actions within the same grade into different sub-grades. The score response associated with action details is first selected from all the score responses by masking redundant information. The selected information, serving as the fine-grained features, is then input into the Fine-Grained Scoring (FGS, see \cref{sec:fgs}) module. Given that the number of sub-grades is typically larger than the number of grades in high-precision assessments, learning an extensive set of sub-grade prototypes might lead to overfitting. To address this concern, the FGS employs a larger set of fixed sub-grade prototypes defined by a simplex Equiangular Tight Frame (ETF) to classify the fine-grained features and predict fine-level sub-grades.

Experimental results demonstrate the significant improvements achieved by CoFInAl compared to state-of-the-art methods with notable gains of 5.49\% and 3.55\% in correlation on two long-term AQA datasets, Rhythmic Gymnastics and Fis-V, respectively. The effectiveness of individual designs is further validated through an extensive ablation study.

Our contributions can be summarized as:
\begin{itemize}
    \item We identify the central challenge of current AQA methods as the domain shift and overfitting, and attribute it to the discrepancy between pre-trained action recognition tasks and fine-tuned AQA tasks.
    \item We propose a coherent coarse-to-fine instruction strategy that aligns AQA tasks with pre-trained models, so as to alleviate the discrepancy.
    \item Our approach can effectively overcome both domain shift and overfitting, and thus achieves substantial improvements across a variety of AQA benchmarks.
\end{itemize}

\section{Related Work}

\myPara{Action Quality Assessment (AQA)} aims to quantitatively evaluate the performance of executed actions across diverse domains \cite{zhou2023hierarchical,zhou2023video,ding2023sedskill,smith-etal-2020-put,wang2020towards}. 
Early methods \cite{pirsiavash2014assessing} heavily relied on handcrafted features, revealing inherent poor generalization \cite{wang2021tsa}. Recent deep learning-based methods \cite{parmar2019and,tang2020uncertainty,yu2021group} have demonstrated improved performance. The primary challenge stems from the relatively small size of existing AQA datasets \cite{wang2021survey}, posing the risk of overfitting. To counter this challenge, pre-trained backbones are commonly employed. \citep{parmar2019and} leveraged C3D \cite{tran2015learning} to enhance AQA performance, \citep{pan2019action} optimized further with I3D \cite{carreira2017quo}, and \citep{xu2022likert} explored the integration of VST \cite{liu2022video} for more powerful feature extraction.  
However, the challenge of domain shift persists as most AQA methods \cite{zhou2023hierarchical,yu2021group} treat the task as a regression problem, with pre-trained backbones invariably trained on action recognition. 
\citep{dadashzadeh2024pecop} proposed a parameter-efficient adapter to address the domain shift to some extent. However, there is still a performance gap due to potential overfitting during the fine-tuning process.
In response to both domain shift and overfitting challenges, we present an innovative solution—\textbf{a novel pre-training alignment for AQA by aligning it with the pre-trained task.}

\myPara{Neural Collapse} describes an elegant geometric structure within the last-layer features and classifier of a well-trained model \cite{papyan2020prevalence}. In a simplified model focusing solely on last-layer optimization, it has been demonstrated as the global optimality in the realm of balanced training with both cross-entropy \cite{graf2021dissecting,fang2021exploring,zhu2021geometric} and MSE \cite{zhou2022optimization,han2021neural,tirer2022extended} loss functions.
Recent investigations have extended the understanding of neural collapse to imbalanced training scenarios, either by fixing a classifier \cite{yang2022we,zhong2023understanding} or introducing novel loss functions \cite{xie2023neural}. Notably, \citep{galanti2021role} have demonstrated that neural collapse remains valid even when transferring a model to new samples or classes. Rather than the previous work \cite{yang2022we,yang2023neural}, we extend an alignment application of the classifier. To the best of our knowledge, \textbf{we are the first to study AQA
from the neural collapse perspective, which offers our method sound interpretability.}

\begin{figure*}
    \centering
    \includegraphics[width=\linewidth,clip,trim=0 42 0 42]{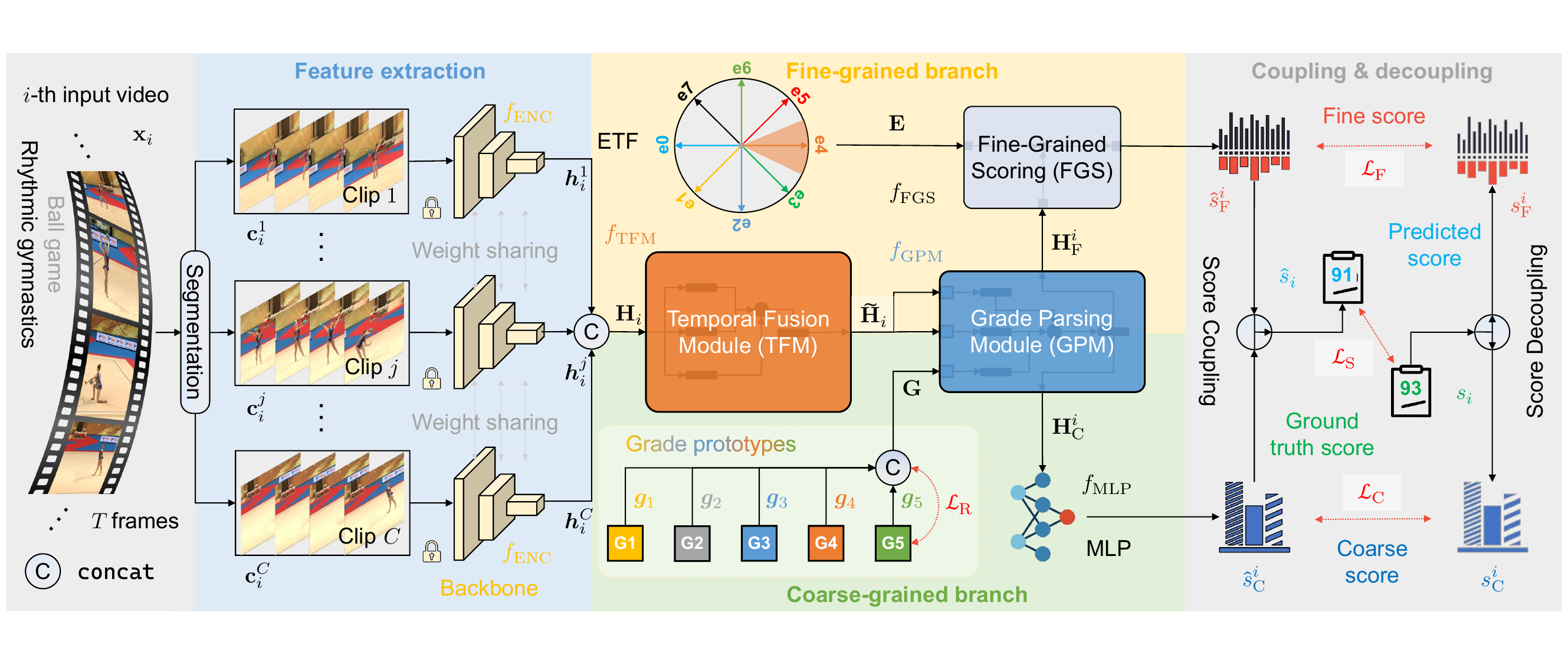}
    \caption{
    CoFInAl Framework: The input video undergoes segmentation into clips for feature extraction using a shared backbone. The Temporal Fusion Module (TFM, see \cref{sec:tfm}) enhances clip features. The Grade Parsing Module (GPM, see \cref{sec:gpm}) then separates features into coarse-grained and fine-grained components. Predictions for coarse-grained and fine-grained scores are derived from these features through an MLP and the Fine-Grained Scoring (FGS, see \cref{sec:fgs}) module. Finally, the final score is coupled with predicted coarse-grained and fine-grained scores. During training, the ground truth score is decoupled to supervise coarse-to-fine learning. 
    }
    \label{fig:framework}
\end{figure*}

\section{Methodology}
In this section, we first introduce the CoFInAl framework, and then elaborate on its core components in detail.

\subsection{Problem Definition and Framework Overview}

\myPara{Vanilla AQA.}
Given a training dataset $\mathcal{D}_{\mathrm{train}} = \{(\mathbf{x}_i, s_i)\}_{i=1}^N$ with $N$ action-score pairs, the goal of AQA is to learn assigning the quality score $s_i \in \mathbb{R}$ to the action $\mathbf{x}_i \in \mathbb{R}^{T \times H \times W \times 3}$ 
with length $T$, resolution $H \times W$, and 3 channels of video frames.
A common practice \cite{parmar2019action} is to employ a pre-trained backbone as the feature extractor $f(\cdot)$ with a regressor $g(\cdot)$ to enhance the performance. This problem can be formulated as follows:

\begin{equation}
\begin{aligned}
    \min_{\theta_f, \theta_g}~& \mathcal{L}_{\mathrm{S}}=\frac{1}{2N}\sum_i(s_i-\hat{s}_i)^2, ~~\\
    \mathrm{s.t.}~&
    \hat{s}_i = g(\bm{h}_i), ~\bm{h}_i = f(\mathbf{x}_i),
\end{aligned}
\end{equation}
where $\bm{h}_i \in \mathbb{R}^{D}$ is the video-level feature, $\hat{s}_i$ is the predicted score, $\mathcal{L}_{\mathrm{S}}$ represents the score regression loss function using MSE, and $\theta_f, \theta_g$ denote the network parameters.

While using pre-trained models from large-scale action recognition datasets improves small-scale AQA performance, domain shift issues persist. Existing methods, exemplified by \citep{zhou2023hierarchical}, often fine-tune pre-trained models for AQA adaptation, risking overfitting and suboptimal outcomes. To address this issue, we introduce CoFInAl, which strategically aligns AQA with the pre-trained task. This approach is inspired by recent advances \cite{zhou2023lima} in large language models that effectively leverage knowledge acquired during pre-training. Specifically, we reformulate AQA as a coarse-to-fine classification problem.

\myPara{Formulation of CoFInAl.} 
Given a type of action, we represent its quality score using $G$ coarse grades to delineate distinct performance levels, with each grade spanning a length of $S_{\mathrm{C}}$. Further, within each grade, we employ $G'$ sub-grades to capture finer variations, and each sub-grade spans a length of $S_{\mathrm{F}}$. Despite the precision error associated with this hierarchical score representation at the fine-grained interval ($S_{\mathrm{F}}$), it allows us to reframe AQA as a classification problem, aligning with the pre-trained task.
Compared to directly dividing the score space into sub-grades, our approach significantly reduces space complexity from $\mathcal{O}(G\times G')$ to $\mathcal{O}(G + G')$, where $G = \lceil S/S_{\mathrm{C}} \rceil$ and $G' = \lceil S_{\mathrm{C}}/S_{\mathrm{F}} \rceil$.

Based on such a system, CoFInAl learns a coarse-to-fine instruction by optimizing the following objective:
\begin{equation} \label{eq:c2f}
    \min_{\Theta} ~~ \mathcal{L} = \mathcal{L}_{\mathrm{S}} + \lambda_{\mathrm{C}} \mathcal{L}_{\mathrm{C}} + \lambda_{\mathrm{F}} \mathcal{L}_{\mathrm{F}} + \lambda_{\mathrm{R}} \mathcal{L}_{\mathrm{R}},
\end{equation}
where $\mathcal{L}_{\mathrm{C}}$ is the coarse-grained loss using the cross-entropy fuction, $\mathcal{L}_{\mathrm{F}}$ is the fine-grained loss (see \cref{eq:rd_loss}), $\mathcal{L}_{\mathrm{R}}$ is the regularization term (see \cref{eq:graph}), $\lambda_{\mathrm{C}}, \lambda_{\mathrm{F}}, \lambda_{\mathrm{R}}$ serve as the loss weights, and $\Theta$ denotes the entire network parameter set.

Next, we provide a brief overview of the entire framework (see \cref{fig:framework}) to offer a high-level understanding of CoFInAl. 

\myPara{Framework Overview.} 
In line with prior work \cite{xu2022likert}, we initially segment the input video $\mathbf{x}_i~(i=1,2,\cdots,N)$ into $C$ clips. To accommodate computational constraints, each clip $\mathbf{c}_i^{j}(j=1,2,\cdots,C)$ undergoes independent processing by the backbone to extract the clip feature $\bm{h}_i^j \in \mathbb{R}^{D_\mathrm{C}}$. Subsequently, these clip features are enhanced to yield the enhanced clip feature $\tilde{\mathbf{H}}_i \in \mathbb{R}^{P\times D_\mathrm{P}}$, as follows:
\begin{equation} \label{eq:tfm}
    \tilde{\mathbf{H}}_i = f_{\mathrm{TFM}}(\mathbf{H}_i),~\mathbf{H}_i=\mathtt{concat}(\bm{h}_i^1,\bm{h}_i^2,\cdots,\bm{h}_i^C),
\end{equation}
where $\mathtt{concat}(\cdot)$ denotes the concatenation operator, and $f_{\mathrm{TFM}}(\cdot)$ indicates the Temporal Fusion Module (TFM, see \cref{sec:tfm}). Next, $\tilde{\mathbf{H}}_i$ is used for coarse-to-fine instruction.

In the initial coarse step, the incorporation of $G$ learnable grade prototypes $\bm{g}_1,\bm{g}_2,\cdots,\bm{g}_G \in \mathbb{R}^{D_\mathrm{P}}$ enables the parsing of enhanced features into two distinct components: coarse-grained and fine-grained features $\mathbf{H}_{\mathrm{C}}^i, \mathbf{H}_{\mathrm{F}}^i \in \mathbb{R}^{G\times D_\mathrm{S}}$, \ie, 
\begin{equation}
    \mathbf{H}_{\mathrm{C}}^i, \mathbf{H}_{\mathrm{F}}^i = f_{\mathrm{GPM}}(\mathbf{G}, \tilde{\mathbf{H}}_i),
\end{equation}
where $\mathbf{G}=\mathtt{concat}(\bm{g}_1,\bm{g}_2,\cdots,\bm{g}_G)$ is the grade prototype matrix and $f_{\mathrm{GPM}}(\cdot)$ indicates the Grade Parsing Module (GPM, see \cref{sec:gpm}). Then, the coarse-grained feature $\mathbf{H}_{\mathrm{C}}^i$ is directly predicted to a grade through an MLP classifier $f_{\mathrm{MLP}}$, \ie, $\hat{s}_{\mathrm{C}}^i = f_{\mathrm{MLP}}(\mathbf{H}_{\mathrm{C}}^i)$, providing an overarching assessment.
In contrast, at the fine step, the utilization of the fine-grained features $\mathbf{H}_{\mathrm{F}}^i$ coupled with a pre-defined ETF matrix facilitates the derivation of the predicted sub-grade $\hat{s}_{\mathrm{F}}^i$, \ie,
\begin{equation} \label{eq:fgs}
    \hat{s}_{\mathrm{F}}^i = f_{\mathrm{FGS}}(\mathbf{H}_{\mathrm{F}}^i,\mathbf{E}),
\end{equation}
where $f_{\mathrm{FGS}}(\cdot)$ represents the Fine-Grained Scoring (FGS, see \cref{sec:fgs}) module. This allows for a more detailed analysis of subtle variations within a specific performance level.

Finally, we can obtain the final predicted score $\hat{s}_i$ by coupling the coarse-grained and fine-grained predictions:
\begin{equation}
    \hat{s}_i = \hat{s}_{\mathrm{C}}^i \times S_{\mathrm{C}} + \hat{s}_{\mathrm{F}}^i \times S_{\mathrm{F}}.
\end{equation}
During training, the ground-truth score $s_i$ is decomposed into the grade $s_{\mathrm{C}}^i=\lfloor s_i /S_{\mathrm{C}} \rfloor$ and the sub-grade $s_{\mathrm{F}}^i= \lfloor(s_i - \lfloor s_i /S_{\mathrm{C}} \rfloor) / S_{\mathrm{F}}\rfloor$ for supervision. This decomposition is instrumental in optimizing the network using \cref{eq:c2f}.

\subsection{CoFInAl: Temporal Fusion Module (TFM)} \label{sec:tfm}
Due to the computational intensity of pre-trained backbones \cite{tran2015learning}, existing AQA methods \cite{yu2021group,zhou2023hierarchical} opt to segment entire video sequences for separate feature extraction. 
However, this compromise poses challenges for a holistic assessment. Actions typically comprise distinct procedures, each associated with key points contributing to the overall assessment. Thus, the initial segmentation may lead to inaccurate evaluations.
To address this, our Temporal Fusion Module (TFM) is designed to strategically fuse temporal cues from individual clips, enabling a more meaningful representation of action procedures.

An inherent challenge in implementing TFM lies in the variability in the number of procedures, which may not always align with the number of clips. To address this, we introduce a procedure-aware self-attention mechanism designed to enhance the representation of individual clips. In the case of an action encompassing $P$ procedures, we initially derive the key, query, and value representations $\mathbf{K}_{\mathrm{T}} \in \mathbb{R}^{C\times D_\mathrm{P}}, \mathbf{Q}_{\mathrm{T}} \in \mathbb{R}^{C\times P \times D_\mathrm{P}}, \mathbf{V}_{\mathrm{T}} \in \mathbb{R}^{C\times D_\mathrm{P}}$ of the initial clip feature $\mathbf{H}_i$ by the linear transformation. Then, we apply average pooling to $\mathbf{Q}_{\mathrm{T}}$ to obtain the new query embedding $\tilde{\mathbf{Q}}_{\mathrm{T}} \in \mathbb{R}^{P \times D_\mathrm{P}}$. This enables us to rewrite \cref{eq:tfm} as follows:
\begin{equation}
    \tilde{\mathbf{H}}_i = \mathtt{softmax}\left(
    \tilde{\mathbf{Q}}_{\mathrm{T}}\mathbf{K}_{\mathrm{T}}^\top \bigg/ \sqrt{D_\mathrm{P}}
    \right) 
 \mathbf{V}_{\mathrm{T}},
\end{equation}
where $\mathtt{softmax}(\cdot)$ denotes the softmax function. 

\myPara{Benefits of TFM.} TFM facilitates the creation of $P$ enhanced clip features, with each clip dedicated to representing a meaningful procedure requiring assessment. In contrast to previous approaches such as the GCN-based method used in \cite{zhou2023hierarchical}, TFM adopts a novel Transformer-based encoder for the fusion process. This novel strategy enhances flexibility and expressiveness in fusing temporal information, thereby contributing to improved action assessment.

\subsection{CoFInAl: Grade Parsing Module (GPM)} \label{sec:gpm}
\begin{figure}
    \centering
    \includegraphics[width=\linewidth,trim=130 75 120 150,clip]{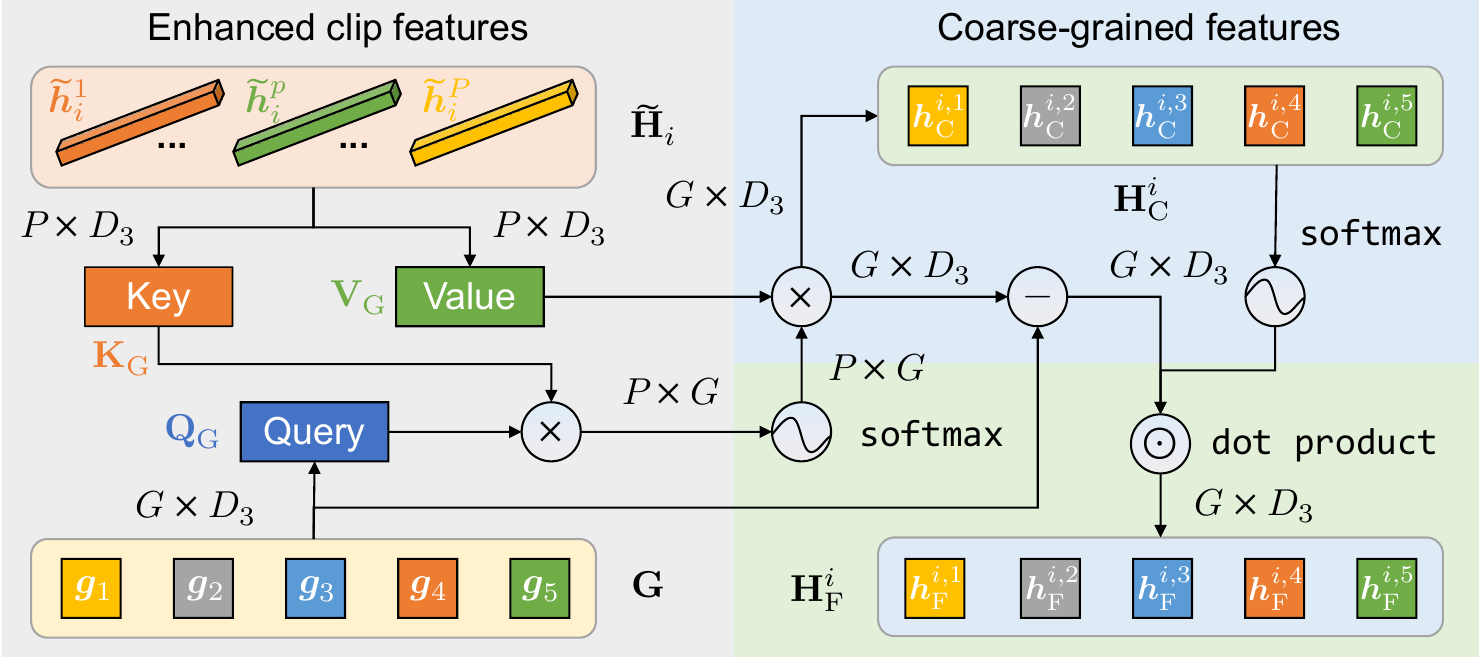}
    \caption{Illustration of Grade Parsing Module (GPM).}
    \label{fig:gpm}
\end{figure}

In sports assessments, judges meticulously analyze various facets of performance, such as precision, creativity, and overall execution. Similarly, our approach acknowledges the inherent complexity of action assessment and aims to replicate a two-step evaluation process. Analogous to a judge first identifying a coarse grade and then discerning variations within each grade, the Grade Parsing Module (GPM, see \cref{fig:gpm}) parses the enhanced features $\tilde{\mathbf{H}}_i$ into coarse-grained and fine-grained components $\mathbf{H}_\mathrm{C}^i,\mathbf{H}_\mathrm{F}^i$, enabling our system to capture both global and detailed performance nuances.

\myPara{Graph Regularization.} To ensure that the network learns assessment facets within the same action type, we introduce $G$ grade prototypes. Each grade prototype encapsulates a standardized rule representation corresponding to a specific performance level, acting as learnable parameters for the network. To guide the learning process and align these grade prototypes with the quality space, we introduce a novel quality-aware graph regularization defined as follows:
\begin{equation} \label{eq:graph}
\mathcal{L}_{\mathrm{R}} = \mathtt{KL}(\arccos( \bar{\mathbf{G}}^\top \bar{\mathbf{G}}, \mathbf{D})),~ \bar{\mathbf{G}} = \mathbf{G} / \|\mathbf{G}\|_2,
\end{equation}
where $\mathtt{KL}(\cdot)$ denotes the Kullback–Leibler (KL) divergence, $\|\cdot\|_2$ represents the $\ell_2$ norm, and $\mathbf{D} \in \mathbb{R}^{G\times G}$ is the distance matrix defined on the quality space with $D_{i,j} = |i-j|$. The regularization ensures that the grade prototypes are anchored in a meaningful space, reducing the risk of overfitting and enhancing the model's generalization capabilities. 

\myPara{Coarse-to-Fine Parsing.} We apply these grade prototypes to query each enhanced clip feature, aggregating the obtained scores to form the coarse-grained feature $\mathbf{H}_\mathrm{C}^i$. Different from TFM, this is implemented by a cross-attention mechanism:
\begin{equation}
    \mathbf{H}_\mathrm{C}^i = \mathtt{softmax}\left(
    \mathbf{Q}_{\mathrm{G}} \mathbf{K}_{\mathrm{G}}^\top \bigg/ \sqrt{D_\mathrm{S}}
    \right) \mathbf{V}_{\mathrm{G}},
\end{equation}
where $\mathbf{Q}_{\mathrm{G}} \in \mathbb{R}^{G \times D_\mathrm{S}}$ is the linear embedding of $\mathbf{G}$, $\mathbf{K}_{\mathrm{G}}, \mathbf{V}_{\mathrm{G}} \in \mathbb{R}^{P \times D_\mathrm{S}}$ are the linear embeddings of $\tilde{\mathbf{H}}_i$. 
Each row in $\mathbf{H}_\mathrm{C}^i$ represents the action's responses across different grades. By examining the distribution of these responses, a judge can make a coarse-level assessment of the action. When a finer evaluation is required, the judge can access the responses associated with the specific grade for further analysis. To execute this procedure, we begin by calculating a mask $\mathbf{M} \in \mathbb{R}^{G \times 1}$, which is:
\begin{equation}
    \mathbf{M} = \mathtt{softmax}(\mathtt{avgPool}(\mathbf{H}_\mathrm{C}^i)),
\end{equation}
where $\mathtt{avgPool}(\cdot)$ denotes the average pooling function along with the feature dimension axis.
To eliminate irrelevant responses, this mask is then applied to the difference between the coarse-grained feature and the grade prototypes $\mathbf{G}$ to obtain the fine-grained feature $\mathbf{H}_\mathrm{F}^i$, which is:
\begin{equation}
\mathbf{H}_\mathrm{F}^i = \mathbf{M} \odot (\mathbf{H}_\mathrm{C}^i - \mathbf{G}),
\end{equation}
where $\odot$ denotes the element-wise dot product operator. 

\myPara{Benefits of GPM.} 
The rationale behind GPM is to replicate the natural two-step evaluation process of judges. This design allows us to reformulate AQA to better align with the pre-trained task. In contrast to the previous work \cite{xu2022likert}, the regularization of grade prototypes aligns the quality-aware space, making it more interpretable. Furthermore, compared with existing methods \cite{yu2021group,zhou2023hierarchical}, the parsing of both coarse and fine features enhances the understanding of global and detailed performance cues, contributing to a holistic action assessment.

\subsection{CoFInAl: Fine-Grained Scoring (FGS)} \label{sec:fgs}
For coarse-grained scoring, we employ a basic MLP classifier. However, the last-layer features of the same class are susceptible to collapsing into their class mean, as indicated by neural collapse theory \cite{papyan2020prevalence}. This collapse tends to eliminate intra-class variations, resulting in inaccurate assessments for AQA. To address this issue, we introduce the Fine-Grained Scoring (FGS) module to maintain such variations within intra-class samples.

Neural collapse theory asserts that the classifier weights on the balanced dataset will converge to a simplex Equiangular Tight Frame (ETF) matrix, defined as:
\begin{definition}[Simplex Equiangular Tight Frame]
	\label{ETF}
	A simplex Equiangular Tight Frame (ETF) refers to a matrix $\mathbf{E}=[\bm{e}_1,\bm{e}_2,\cdots,\bm{e}_K]\in\mathbb{R}^{d \times K}$, which satisfies:
	\begin{equation}\label{ETF_M}
		\mathbf{E} =\sqrt{\frac{K}{K-1}}\mathbf{U}\left(\mathbf{I}_K-\frac{1}{K}\mathbf{1}_K\mathbf{1}_K^\top\right),
	\end{equation}
	where $\mathbf{I}_K \in \mathbb{R}^{K \times K}$ is the identity matrix, $\mathbf{1}_K \in \mathbb{R}^{K\times 1}$ is an all-ones vector, and $\mathbf{U}\in\mathbb{R}^{d\times K}$ allows a rotation and satisfies $\mathbf{U}^\top\mathbf{U}=\mathbf{I}_K$.  All column vectors in $\mathbf{E}$ have the same $\ell_2$ norm and any pair has an inner product of $-\frac{1}{K-1}$, \emph{i.e.,}
\begin{equation}\label{mimj}
	\bm{e}^\top_{i}\bm{e}_{j}=\frac{K}{K-1}\delta_{ij}-\frac{1}{K-1},\ \ \forall i, j \in \{1,2, \cdots, K\},
\end{equation}
where $\delta_{ij}=0$ when $i\neq j$, and 1 otherwise. 
\end{definition}

In contrast to coarse-grained scoring, our FGS module leverages a non-learnable ETF matrix $\mathbf{E}$ with dimensions $G' \times D_\mathrm{S}$. The key advantages of this approach are twofold: (1) The non-learnable nature of ETF mitigates overfitting, especially considering that the number of fine-grained levels typically exceeds that of coarse-grained levels. (2) The predefined matrix embodies the optimality of a classifier. In this way, the model prediction can be simplified to the nearest class centers, \ie, \cref{eq:fgs} can be rewritten as follows:
\begin{equation}
    \hat{s}_{\mathrm{F}}^i = \mathop{\arg\max}_j \langle\bm{h}_\mathrm{F}^i, \bm{e}_j \rangle, ~ \bm{h}_\mathrm{F}^i = \mathtt{avgPool}(\mathbf{H}_\mathrm{F}^i),
\end{equation}
where $\langle\cdot\rangle$ denotes the inner-product operator.
Accordingly, the fine-grained loss $\mathcal{L}_{\mathrm{F}}$ in \cref{eq:c2f} can be defined as follows:
\begin{equation} \label{eq:rd_loss}
    \mathcal{L}_{\mathrm{F}} = \frac{1}{2N} \sum_i \left( \langle \bar{\bm{h}}_{\mathrm{F}}^i, \hat{\bm{e}}_i \rangle -1 \right)^2,
\end{equation}
where $\bar{\bm{h}}_\mathrm{F}^i = \bm{h}_\mathrm{F}^i/\|\bm{h}_\mathrm{F}^i\|_2$ and $\hat{\bm{e}}_i$ denotes the predicted fixed prototype in $\mathbf{E}$ with respect to the sub-grade $\hat{s}_{\mathrm{F}}^i$.


\myPara{Benefits of FGS.} The novel application of the ETF matrix in our FGS module ensures capturing fine-grained details effectively, enhancing the precision of our framework in discerning subtle variations. Importantly, in addressing domain shift, the non-learnable nature of ETF aligns with the pre-trained task, mitigating overfitting issues associated with shifts in data distribution. This strategic alignment contributes to the robustness of our model, setting it apart from existing AQA methods \cite{yu2021group,zhou2023hierarchical} and making it well-suited for real-world applications with diverse datasets.

\section{Experiments}
In this section, we first describe the experimental setup, and then present and analyze the experimental results.

\begin{table*}
    \centering
    \resizebox{\linewidth}{!}{
    \begin{tabular}{rlcccccccc}
    \toprule
    \multirow{2}{*}[-0.5ex]{Method} & \multirow{2}{*}[-0.5ex]{Backbone} & \multicolumn{5}{c}{RG} & \multicolumn{3}{c}{Fis-V} \\ \cmidrule(lr){3-7}  \cmidrule(lr){8-10}
    & & Ball & Clubs & Hoop & Ribbon & Average & TES & PCS & Average \\
    \midrule
    C3D+SVR \cite{parmar2017learning} & C3D & 0.357 & 0.551 & 0.495 & 0.516 & 0.483 & 0.400 & 0.590 & 0.501 \\
    MS-LSTM \cite{xu2019learning} & C3D & - & - & - & - & - & 0.650 & 0.780 & 0.721 \\
    \hline
    MS-LSTM \cite{xu2019learning} & I3D & 0.515 & 0.621 & 0.540 & 0.522 & 0.551 & - & - & -\\
    ACTION-NET \cite{zeng2020hybrid} & I3D + ResNet & 0.528 & 0.652 & 0.708 & 0.578 & 0.623 & - & - & - \\
    GDLT \cite{xu2022likert}  & I3D & 0.526 & \underline{0.710} & \underline{0.729} & \underline{0.704} & \underline{0.674} & 0.260 & 0.395 & 0.329 \\
    HGCN$^\star$ \cite{zhou2023hierarchical} & I3D  & \underline{0.534} & 0.609 & 0.706 & 0.621 & 0.621 & \underline{0.311} & \underline{0.407}  & \underline{0.360}  \\
  
    \rowcolor{LightCyan} CoFInAl (Ours) & I3D & \textbf{0.625} &   \textbf{0.719} & \textbf{0.734} & \textbf{0.757} &\textbf{0.712}  &  \textbf{0.589} & \textbf{0.788}  &  \textbf{0.702} \\
    \hline
    
    MS-LSTM \cite{xu2019learning} & VST & 0.621 & 0.661 & 0.670 & 0.695 & 0.663 & 0.660 & 0.809 & 0.744 \\
    ACTION-NET \cite{zeng2020hybrid} & VST + ResNet & 0.684 & 0.737 & 0.733 & \underline{0.754} & 0.728 & \underline{0.694} & 0.809 & 0.757 \\
    GDLT \cite{xu2022likert}  & VST & \underline{0.746} & \underline{0.802} & \underline{0.765} & 0.741 & \underline{0.765} & 0.685 & \underline{0.820} & \underline{0.761} \\
    HGCN$^\star$ \cite{zhou2023hierarchical} & VST & 0.711 & 0.789 & 0.728 & 0.703  & 0.735 & 0.246 & 0.221 & 0.234 \\
    \rowcolor{LightCyan} CoFInAl (Ours) & VST & \textbf{0.809} & \textbf{0.806} & \textbf{0.804} & \textbf{0.810} & \textbf{0.807} & \textbf{0.716} &  \textbf{0.843} & \textbf{0.788} \\
    \bottomrule
    \end{tabular}
    }
    \caption{
    Experimental results of SRCC on RG and Fis-V datasets. 
    The best results are presented in \textbf{bold}, while the second-best results are \underline{underlined}. 
    The symbol $\star$ denotes our reimplementation based on the official code.
    The average SRCC is calculated using the Fisher-z value.
    }
    \label{tab:sota}
    \vspace{-0.2cm}
\end{table*}

\subsection{Experimental Setups}
\myPara{Datasets.} 
In this work, we evaluate all models on two comprehensive long-term AQA datasets. The \textbf{Rhythmic Gymnastics (RG)} dataset \cite{zeng2020hybrid} comprises a total of 1000 videos featuring four distinct rhythmic gymnastics actions performed with various apparatuses, including ball, clubs, hoop, and ribbon. Each video has an approximate duration of 1.6 minutes, and the frame rate is set at 25 frames per second. The dataset is divided into training and evaluation sets, with 200 videos allocated for training and 50 for evaluation in each action category.
The \textbf{Figure Skating Video (Fis-V)} dataset \cite{pirsiavash2014assessing,parmar2017learning} consists of 500 videos capturing ladies' singles short programs in figure skating. Each video has a duration of approximately 2.9 minutes, and the frame rate is set at 25 frames per second. Adhering to the official split, the dataset is divided into 400 training videos and 100 testing videos. All videos come with annotations for two scores: Total Element Score (TES) and Total Program Component Score (PCS). Following \citep{xu2019learning}, we develop independent models for different score/action types.

\myPara{Metric.}
Consistent with previous studies \cite{yu2021group,xu2022likert}, we utilize the Spearman's Rank Correlation Coefficient (SRCC) as the evaluation metric, denoted as $\rho$. The SRCC is defined as the Pearson correlation coefficient between two rank vectors, $\bm{p}$ and $\bm{q}$, with respect to predicted and ground-truth scores, which can be formulated as follows:
\begin{equation}
  \rho = \frac{ \sum_i (p_i - \bar{p}) (q_i - \bar{q}) }{\sqrt{\sum_i (p_i - \bar{p})^2 \sum_i (q_i - \bar{q})^2}},
\end{equation}
where $\bar{p}$ and $\bar{q}$ denote the average values of the rank vectors $\bm{p}$ and $\bm{q}$, respectively. A higher SRCC indicates a stronger rank correlation between predicted and ground-truth scores.
Following \citep{pan2019action}, we compute the average SRCC across different action types for RG and score types for Fis-V by aggregating individual SRCCs using the Fisher’s z-value.

\myPara{Implementation Details.} 
CoFInAl is implemented using PyTorch on a GPU for efficient parallel processing. We employ VST pre-trained on Kinetics 600 \cite{xu2022likert} as the backbone and fixed for AQA. The feature dimensions $D_\mathrm{C},D_\mathrm{P},D_\mathrm{S}$ are set to 1024, 512, and 256, respectively. 
We initially partition the video into non-overlapping 32-frame segments. During training, we randomly determine the start segment, specifically $C=68$ for RG and $C=124$ for Fis-V. During testing, all segments are utilized.
The number of procedures $P$ is fixed at 5 for all actions. We optimize all models using SGD with a momentum of 0.9. The batch size is 32, and the learning rate starts at 0.01, gradually decreasing to 0.0001 through a cosine annealing strategy. The number of epochs is set to 200. The loss weights $\lambda_{\mathrm{C}},\lambda_{\mathrm{F}},\lambda_{\mathrm{R}}$ are set to 1. To further regularize the models, we apply a dropout of 0.3/0.7 for RG/Fis-V, and the weight decay is set to 0.01. 

\subsection{Results and Analysis}
We present the primary experiments. For more experimental results, we direct readers to the supplementary material.

\myPara{Comparisons with the State-of-the-Art.}
We conduct a comparison of various state-of-the-art methods, including C3D+SVR \cite{parmar2017learning}, MS-LSTM \cite{xu2019learning}, ACTION-NET \cite{zeng2020hybrid}, and GDLT \cite{xu2022likert}. The results are reported in \cref{tab:sota}. 

To demonstrate the effect of different backbones, \cref{tab:sota} compares diverse architectures including C3D \cite{tran2015learning}, I3D \cite{carreira2017quo}, ResNet \cite{he2016deep}, and VST \cite{liu2022video}. I3D consistently outperforms C3D on RG, showcasing its superior ability to capture temporal dynamics. VST stands out as the most effective backbone, achieving the highest average SRCC on RG and Fis-V. 
For instance, the results of ACTION-NET with the VST backbone on RG showcase a notable correlation gain of over 16.85\% compared to its counterpart with I3D. This underscores that VST outperforms I3D in capturing the subtle temporal dynamics crucial for AQA. Particularly, CoFInAl with I3D and VST achieves excellent results in all categories, while CoFInAl with VST achieves the best results, emphasizing the effectiveness of the proposed alignment framework and the superiority of VST. These findings underscore the significance of advanced backbones in capturing detailed action details for improved AQA performance.

CoFInAl with VST consistently outperforms others across all actions and score types in both datasets, demonstrating its effectiveness in aligning AQA with pre-trained tasks. Notably, it excels in categories like Ball, Hoop, and Ribbon on RG and across TES and PCS on Fis-V by a large margin. For example, CoFInAl achieves a remarkable 8.45\% correlation gain in Ball compared to the second-best GDLT. Overall, CoFInAl delivers significant correlation gains of 5.49\% and 3.55\% on RG and Fis-V, respectively. These results underscore the benefits of CoFInAl in addressing domain shift challenges. The comparison in \cref{tab:sota} positions CoFInAl as a promising solution for advancing AQA fields.

\begin{table}
    \centering
    \resizebox{\linewidth}{!}{
    \begin{tabular}{rcccc}
    \toprule
    Method &  \makecell{FLOPs \\ (G)} &  \makecell{Parameter \\ (M)} & \makecell{Inference Time \\ (ms)}   \\
    \midrule
    ACTION-NET  \cite{zeng2020hybrid}  & 34.7500 & 28.08 &  305.2474 \\
    GDLT  \cite{xu2022likert}  & ~~0.1164 & ~~3.20  & ~~~~3.2249 \\
    HGCN  \cite{zhou2023hierarchical} & ~~1.1201 & ~~0.50 & ~~~~6.7830\\
    \rowcolor{LightCyan} CoFInAl (Ours)  & ~~0.1178 & ~~3.70 & ~~~~3.8834\\
    \bottomrule
    \end{tabular}
    }
    \caption{Computational comparison with existing methods.}
    \label{tab:computation}
\end{table}

We have tested the model efficiency under identical conditions. The results in \cref{tab:computation} highlight CoFInAl's significant computational efficiency compared to ACTION-NET and HGCN. Notably, our method builds upon GDLT, with a mere 0.0014G computation expansion, 0.50M parameters expenditure, and 0.6585ms inference delay. This can be attributed to coarse-to-fine alignment, which enhances performance with minimal computational overhead. 

\myPara{Ablation Study.}
We conduct a comprehensive ablation study to explore the individual contributions of core components within CoFInAl, presenting the results in \cref{tab:ablation}. Each row in the table corresponds to a specific configuration, and the columns represent different action types on the RG dataset.

\begin{table}[]
    \centering
    \small
    \resizebox{\linewidth}{!}{
    \begin{tabular}{llllll}
    \toprule
    Setting & Ball & Clubs & Hoop & Ribbon \\
    \midrule
    Ours  & 0.809 & 0.806 & 0.804 & 0.810  \\
    \hline
    ~~w/o $\mathcal{L}_{\mathrm{R}}$ & 0.777 $^{\downarrow 4\%}$ & 0.746 $^{\downarrow 7\%}$  & 0.786 $^{\downarrow 4\%}$ & 0.750 $^{\downarrow 7\%}$ \\
    ~~w/o TFM & 0.590 $^{\downarrow 27\%}$ & 0.705 $^{\downarrow 13\%}$ & 0.685 $^{\downarrow 15\%}$ & 0.761 $^{\downarrow 6\%}$  \\
    ~~w/o GPM & 0.332 $^{\downarrow 59\%}$ & 0.511 $^{\downarrow 36\%}$ & 0.054 $^{\downarrow 93\%}$ &  0.493 $^{\downarrow 39\%}$  \\
    ~~w/o FGS  & 0.702 $^{\downarrow 13\%}$  & 0.754 $^{\downarrow 6\%}$ & 0.709 $^{\downarrow 12\%}$ & 0.757 $^{\downarrow 7\%}$   \\
    \bottomrule
    \end{tabular}
    }
    \caption{Ablation results on the RG dataset.}
    \vspace{-0.5cm}
    \label{tab:ablation}
\end{table}

The removal of the regularization term ($\mathcal{L}_{\mathrm{R}}$) results in a slight decrease in performance across all actions.
This indicates the importance of regularizing the grades to align with the quality-aware space, especially in the context of AQA with limited labeled data. Removing TFM leads to a substantial performance drop, especially in Ball and Hoop, with 27\% and 15\% correlation drops, respectively. This underscores the importance of capturing temporal dependencies in AQA.
The absence of GPM results in severe performance degradation across all actions, with a dramatic 93\% correlation drop in Hoop. This emphasizes the necessity of hierarchical processing for effective AQA, aligning with the judging process of providing coarse and fine-grained assessments.
Removing FGS leads to a notable decrease in performance, especially in Ball and Hoop, with 13\% and 12\% correlation drops, respectively. This highlights the significance of discerning variations within grades for nuanced AQA.

\begin{figure}
    \centering
    \includegraphics[width=\linewidth]{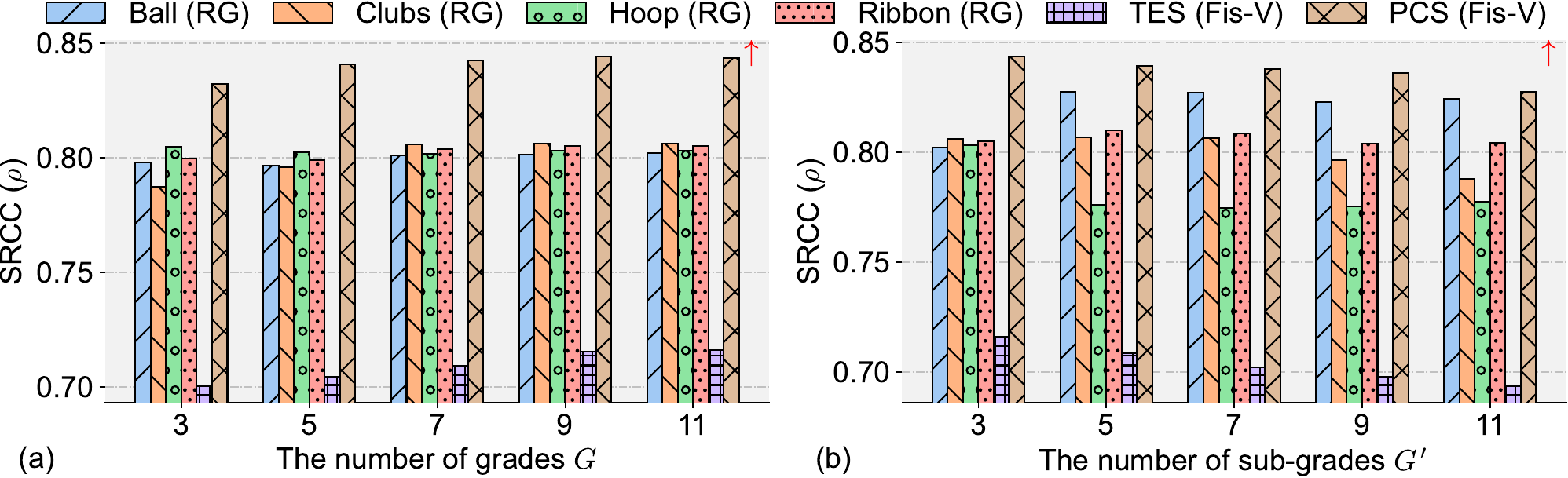}
    \caption{SRCC bars of the number of (a) grades and (b) sub-grades.}
    \label{fig:bars}
\end{figure}

\myPara{Effectiveness of Grades and Sub-Grades.}
We systematically varied the number of grades and sub-grades to identify the optimal parameter combination. The results in \cref{fig:bars} indicate that excessively large or small numbers of grades lead to suboptimal outcomes. For instance, when the number of grades is set to 7, the best results are achieved in Clubs (RG). This outcome is attributed to the judicious balance achieved with four grades, ensuring a nuanced yet comprehensible assessment. Similar observations hold for sub-grades, reinforcing the importance of carefully selecting the number of grades and sub-grades for effective action assessment.

\begin{figure}
    \centering
    \includegraphics[width=\linewidth]{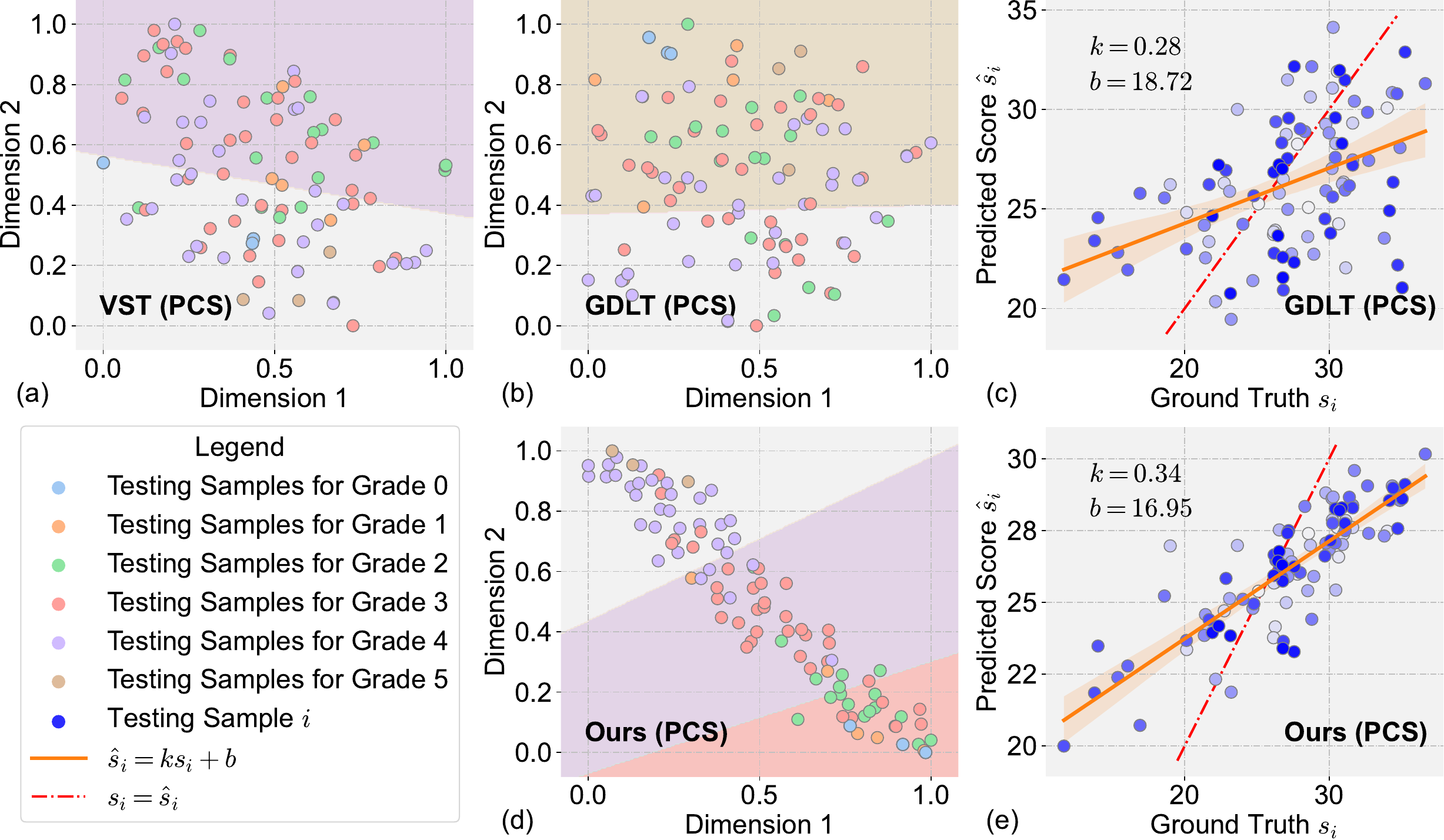}
    \caption{T-SNE feature distribution plots (a, b, d) and correlation comparison plots (c, e) contrasting GDLT with our CoFInAl method.}
    \label{fig:tsne}
\end{figure}

\myPara{Effectiveness in Addressing Domain Shifts.}
We evaluate the effectiveness of our CoFInAl method in addressing domain shifts by visualizing the feature distribution in the latent space and comparing correlation plots between the predicted and ground truth scores. Leveraging the T-SNE toolbox \cite{van2008visualizing}, we present feature and correlation plots in \cref{fig:tsne}, which are conducted on PCS (Fis-V). For feature distribution plots, an SVC classifier is applied to segment the plane into different areas, enhancing visualization. Samples of the same grade (in the same color shading) occupying the same area indicate better feature distribution for action assessment. Additionally, we include a comparison with GDLT \cite{xu2022likert} in \cref{fig:tsne}.

Initially, we categorized samples in the entire PCS testing set into six grades, assigning grade labels from 0 to 5. Specifically, \cref{fig:tsne}(a) illustrates features extracted by the VST backbone, revealing a mixed sample distribution that poses challenges in distinguishment. This highlights the suboptimal suitability of the pre-trained backbone for the AQA task due to domain shifts from broader tasks to AQA. In \cref{fig:tsne}(b), we observe the feature distribution of GDLT, which appears confused, indicating inappropriate feature learning. In contrast, \cref{fig:tsne}(d) demonstrates the linear feature distribution of our CoFInAl, facilitating clearer identification of samples in each grade. This underscores the superiority of CoFInAl over existing methods, attributed to the coarse-to-fine instruction aligning AQA with broader pre-trained tasks. Finally, we compare correlation plots between GDLT and CoFInAl. The regressed correlation line ($\hat{s}_i=k s_i + b$) of CoFInAl (see \cref{fig:tsne}(e)) is closer to the ideal line ($\hat{s}_i=s_i$) than that of GDLT (see \cref{fig:tsne}(c)), which indicates the higher correlation of CoFInAl and thus further emphasizes its effectiveness.

\begin{figure}
    \centering
    \includegraphics[width=\linewidth]{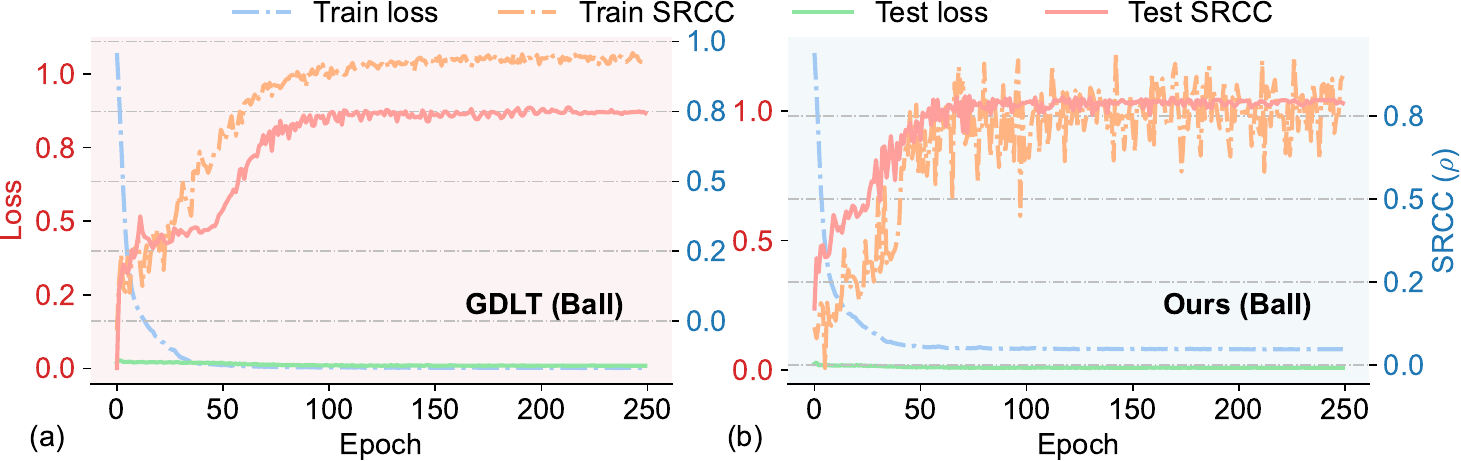}
    \caption{Comparison of loss and SRCC with GDLT on Ball (RG).}
    \label{fig:loss}
\end{figure}

\myPara{Effectiveness in Mitigating Overfitting.}
\cref{fig:loss} provides a performance comparison in terms of loss and SRCC between GDLT \cite{xu2022likert} (see \cref{fig:loss}(a)) and our proposed method (see \cref{fig:loss}(b)) on the Ball (RG) dataset. In \cref{fig:loss}(a), it is observed that during training, the training SRCC surpasses the testing SRCC after approximately 25 epochs, and the gap continues to widen, indicating a severe overfitting issue with GDLT. In contrast, the testing SRCC in \cref{fig:loss}(b) remains higher than the training SRCC, suggesting that our method avoids overfitting. These results in \cref{fig:loss} underscore the effectiveness of our approach in mitigating overfitting issues compared to the state-of-the-art GDLT.

\section{Conclusion}
In conclusion, we introduce CoFInAl as a solution to address the key challenges of AQA. Specifically, CoFInAl integrates a two-step learning process: coarse-grained grade prototype learning and fine-grained sub-grade classification using the simplex ETF. By aligning AQA with broader pre-trained tasks, CoFInAl effectively navigates the issues of domain shift and overfitting. Experiments on RG and Fis-V demonstrate the effectiveness of CoFInAl. Its ability to effectively align downstream tasks with pre-training objectives opens avenues for enhanced model generalization and performance in broader machine learning contexts.

\myPara{Potential Limitations and Future Work.}
The effectiveness of CoFInAl relies on the assumed transferability between pre-trained tasks and AQA. 
Challenges may arise in specialized scenarios where conflicts between pre-trained features and AQA cues impact performance.
Future work should explore advanced adaptation strategies to better capture AQA cues. 



\section*{Acknowledgments}
This work was supported in part by the National Natural Science Foundation of China under Project 62272019 (Liang), and also in part by the International Joint Doctoral Education Fund of Beihang University (Zhou).

\bibliographystyle{named}
\bibliography{aqa}

\input{supp_content}

\end{document}

%% file: supp_content.tex
\renewcommand{\theequation}{\textcolor{gray}{S}\arabic{equation}}
\renewcommand{\thetable}{\textcolor{gray}{S}\arabic{table}}
\renewcommand{\thefigure}{\textcolor{gray}{S}\arabic{figure}}
\renewcommand{\thepage}{\textcolor{gray}{S}\arabic{page}}
\appendix

\newpage

\section{Additional Implementation Details}

\myPara{Feature Alignment} is a key technique that enables CoFInAl to translate between the pre-training and fine-tuning domains effectively. Similar to recent advances in large language model tuning, aligning objectives facilitates reconstructing the target task to be more consistent with the pre-trained model's capabilities.
By projecting features into a standardized space, task-specific nuances can be mapped to better align with the model's conceptual understanding derived from pre-training. For instance, GDLT employs feature alignment to map into the range $[0, D_{\mathrm{C}}]$, where $D_{\mathrm{C}}$ denotes the feature dimension.

However, as a complete feature space may encompass complex geometries spanning various domains, restricting alignment to a quadrant space can be limiting. CoFInAl introduces alignment methods that can normalize features into bounded ranges like $[0,1]$ and symmetrical spaces like $[-1, 1]$ that are more representative.
Comparative experiments discussed in \cref{sec_exp} further analyze the impact of different alignment ranges as hyperparameters. The results demonstrate the flexibility of CoFInAl in translating tasks through principled feature alignment while preserving their nuanced intricacies. The meanings of different alignment methods are shown in \cref{tab:align}.
\begin{table}[h]
    \small
    \centering
    \begin{tabular}{cccc}
    \toprule
    Alignment Mode   & 0 & 1 & 2  \\
    \midrule
    Boundary Range     &  $[0, D_{\mathrm{C}}]$ & $[-1, 1]$ & $[0,1]$  \\
    \bottomrule
    \end{tabular}
    \caption{Meanings of different feature space alignment methods.}
    \label{tab:align}
\end{table}

\begin{figure}[h]
    \centering
    \includegraphics[width=\linewidth]{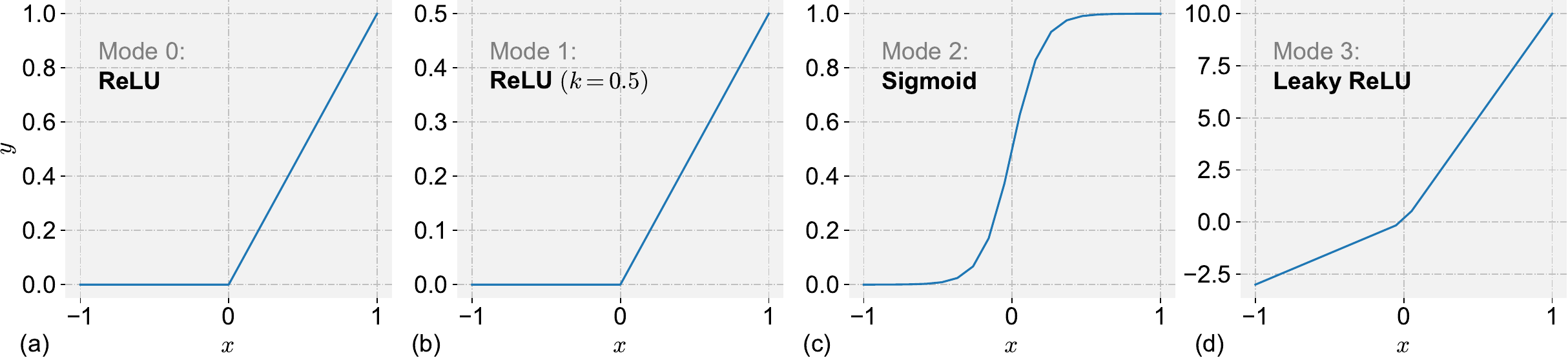}
    \caption{Visualization of different activation functions.}
    \label{fig:act}
\end{figure}

\myPara{Activation.} The choice of the activation function for the final layer transforming features into grade/score predictions can significantly impact model convergence and effectiveness. We conduct comparative studies evaluating four activation options, as shown in \cref{fig:act}. The results provide insight into the tradeoffs. For instance, the probabilistic range from {\tt Sigmoid} leads to faster early convergence compared to unconstrained {\tt RELU}, at the cost of precision. {\tt RELU(k = 0.5)} strikes a balance in stabilizing training.

\myPara{Others.} 
To enable accurate reproduction of reported results under the same hyperparameter settings, the random seed for weight initialization and dataset shuffling is fixed at 0 across all experiments. Setting the global random seed eliminates variance across multiple trials and provides a fair basis for comparing model performance. 
The score predictor in CoFInAl, consisting of a simple MLP or ETF-based classifier, is relatively shallow compared to the convolutional feature extractor.
By randomly initializing the scorer, CoFInAl constructs a prediction function purely based on the training data distribution without inherited biases. This allows jointly learning specialized score alignment heuristics attuned to the target task metrics, unconstrained by external assumptions.

\section{Additional Results and Analysis} \label{sec_exp}

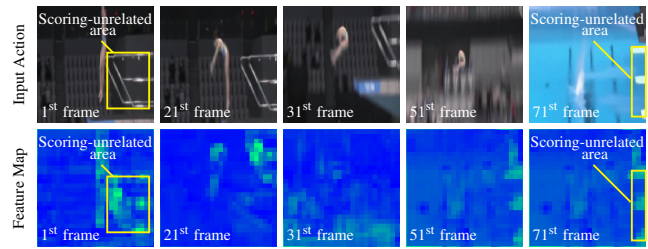
\begin{figure}[h]
    \centering
    \setlength{\tabcolsep}{0.1em}
    \begin{tabular}{m{0.03\linewidth}<{\centering}m{0.95\linewidth}<{\centering}}
    \rotatebox{90}{\tiny Input Action}    &  
    \begin{overpic}[width=0.190\linewidth]{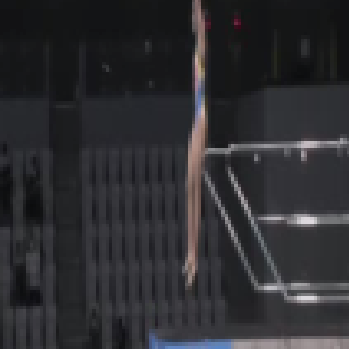}
        \put(3,5){\tiny\color{white}1$^{\text{st}}$ frame}
        \put(60,12){\linethickness{0.25mm}\color{yellow}\polygon(0,0)(36,0)(36,48)(0,48)}
        \put(55,70){\linethickness{0.25mm}\color{yellow}\line(1,-1){10}}
        \put(1,85){\tiny\color{white} Scoring-unrelated area}
        \put(45,75){\tiny\color{white} area}
    \end{overpic}
    \begin{overpic}[width=0.190\linewidth]{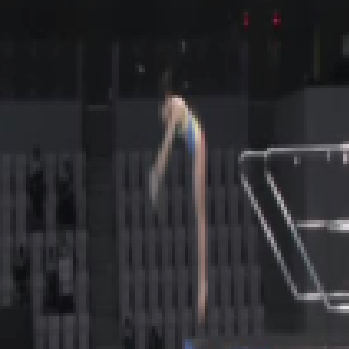}
        \put(3,5){\tiny\color{white}21$^{\text{st}}$ frame}
    \end{overpic}
    \begin{overpic}[width=0.190\linewidth]{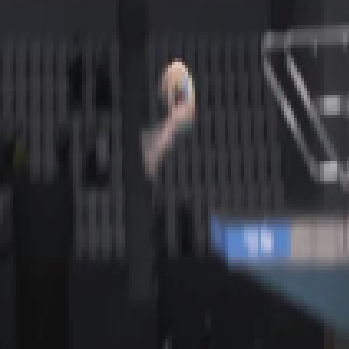}
        \put(3,5){\tiny\color{white}31$^{\text{st}}$ frame}
    \end{overpic}
    \begin{overpic}[width=0.190\linewidth]{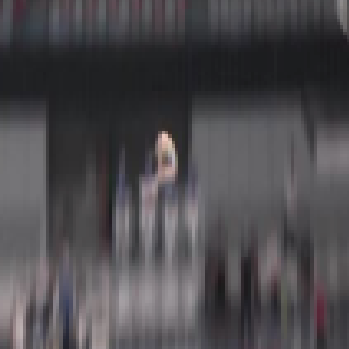}
        \put(3,5){\tiny\color{white}51$^{\text{st}}$ frame}
    \end{overpic}
    \begin{overpic}[width=0.190\linewidth]{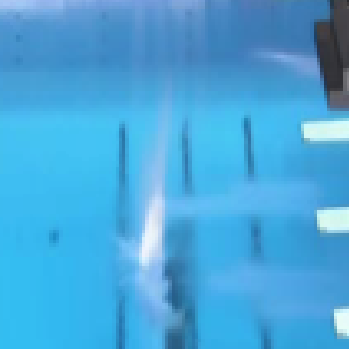}
        \put(3,5){\tiny\color{white}71$^{\text{st}}$ frame}
        \put(88,5){\linethickness{0.25mm}\color{yellow}\polygon(0,0)(12,0)(12,60)(0,60)}
        \put(55,70){\linethickness{0.25mm}\color{yellow}\line(1,-1){32}}
        \put(1,85){\tiny\color{white} Scoring-unrelated area}
        \put(45,75){\tiny\color{white} area}
    \end{overpic} \\
    \rotatebox{90}{\tiny Feature Map}     & 
    \begin{overpic}[width=0.190\linewidth]{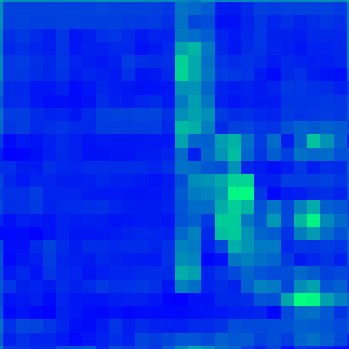}
        \put(3,5){\tiny\color{white}1$^{\text{st}}$ frame}
        \put(60,12){\linethickness{0.25mm}\color{yellow}\polygon(0,0)(36,0)(36,48)(0,48)}
        \put(55,70){\linethickness{0.25mm}\color{yellow}\line(1,-1){10}}
        \put(1,85){\tiny\color{white} Scoring-unrelated area}
        \put(45,75){\tiny\color{white} area}
    \end{overpic}
    \begin{overpic}[width=0.190\linewidth]{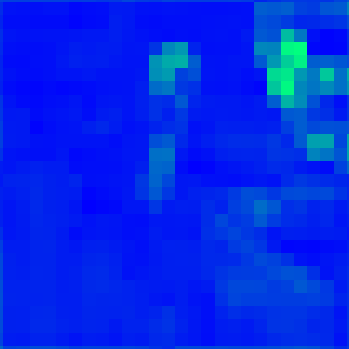}
        \put(3,5){\tiny\color{white}21$^{\text{st}}$ frame}
    \end{overpic}
    \begin{overpic}[width=0.190\linewidth]{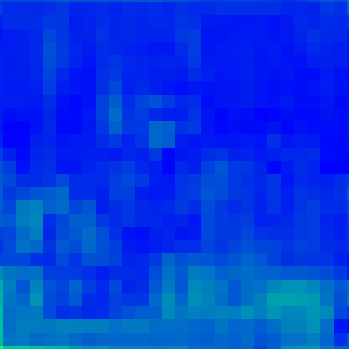}
        \put(3,5){\tiny\color{white}31$^{\text{st}}$ frame}
    \end{overpic}
    \begin{overpic}[width=0.190\linewidth]{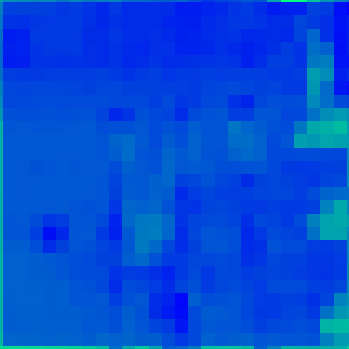}
        \put(3,5){\tiny\color{white}51$^{\text{st}}$ frame}
    \end{overpic}
    \begin{overpic}[width=0.190\linewidth]{figs/06-06-feature_map}
        \put(3,5){\tiny\color{white}71$^{\text{st}}$ frame}
        \put(88,5){\linethickness{0.25mm}\color{yellow}\polygon(0,0)(12,0)(12,60)(0,60)}
        \put(55,70){\linethickness{0.25mm}\color{yellow}\line(1,-1){32}}
        \put(1,85){\tiny\color{white} Scoring-unrelated area}
        \put(45,75){\tiny\color{white} area}
    \end{overpic}
    \end{tabular}
    
    \caption{Visualization of domain shift from action recognition to AQA: The pre-trained I3D backbone emphasizes coarse features (highlighted in yellow boxes), potentially unrelated to scoring.}
    \label{fig:ds}
\end{figure}

\textbf{Domain Shift Visualization.}
In \cref{fig:ds}, we provide a feature map visualization of a diving sample to demonstrate the domain shift differently. The pre-trained backbone highlights features such as handrails and diving boards (highlighted in yellow boxes), crucial for action recognition but potentially unrelated to scoring. This can primarily be attributed to the broader focus of the pre-trained task on coarse-level features.

\begin{figure}
    \centering
    \includegraphics[width=\linewidth]{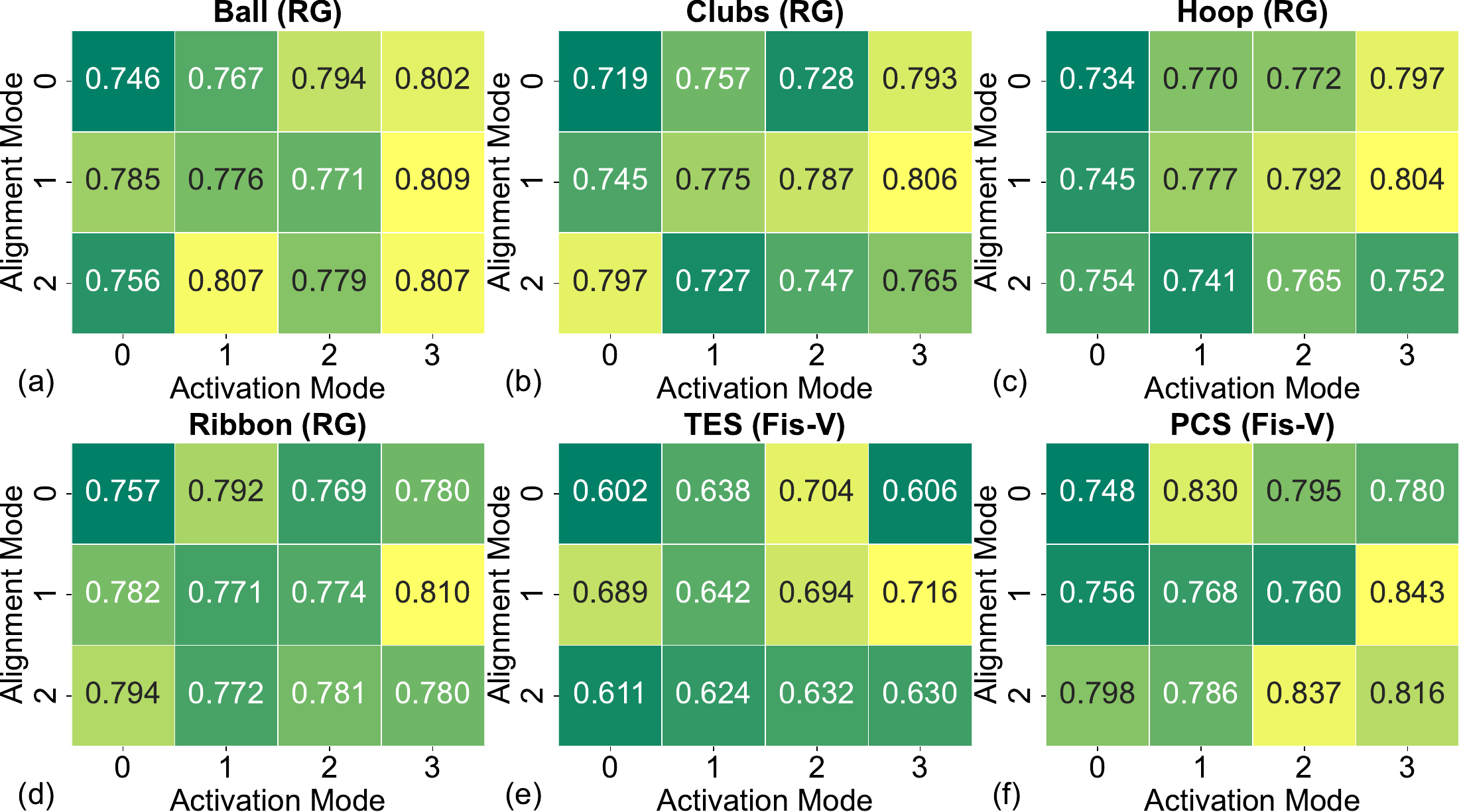}
    \caption{SRCC heatmaps of different alignment and activation modes on the RG and Fis-V datasets.}
    \label{fig:config}
\end{figure}

\myPara{Sensitivity Analysis of Hyperparameters.}
The hyperparameter analysis in \cref{fig:config} provides valuable insights into optimizing the alignment of CoFInAl. Specifically, the experiments evaluate different feature alignment ranges and prediction activation functions across diverse AQA categories.
The heatmaps showcase certain consistent trends. Feature alignment to the range $[-1, 1]$ (Mode 1) generally outperforms simple non-negative mapping (Mode 0), indicating the benefits of accounting for bidirectional deviations. Meanwhile, the {\tt Leaky ReLU} activation (Mode 3) offers a performance boost over others, demonstrating the importance of stabilized training.
Overall, the enriching results advocate for a comprehensive, granular investigation of task alignment techniques across diverse modalities. Principled experimentation to unravel geometric intricacies can unlock generalized alignment heuristics for versatile score translation.

\myPara{Effectiveness of the Number of Procedures.}
In our study, ``procedures" means key segments or phases relevant to the specific actions. 
We have considered several points:
(1) Ideally, these procedures should be identified based on domain-specific knowledge. The number of procedures can vary depending on the sport or activity being analyzed. 
(2) However, collecting such annotations can be expensive. Without this information, the network may struggle to correlate procedures with meaningful key parts such as jumps. 
(3) Instead, these procedures represent optimal segments of actions, with each segment contributing to the final score. This segmentation is necessary as scoring long-term actions with the entire video at once is challenging.
(4) In reality, we conducted a pilot test to obtain the optimal hyperparameter by experimenting with different numbers of procedures. The results in \cref{tab:p_ablation} indicate setting $P$ to 5 yields the best performance. 

\begin{table}[h]
    \centering
    \resizebox{\linewidth}{!}{
    \begin{tabular}{cccccccc}
    \toprule
    $P$  &    1  &  2  &  3 &  4  &  5 & 6 & 7 \\
    \midrule
    SRCC   & 0.819 & 0.821 &  0.808 &  0.810 & \textbf{0.848} & 0.813 & 0.807 \\
    \bottomrule
    \end{tabular}
    }
    \caption{Pilot test for $P$ on the Fis-V (PCS) dataset.}
    \label{tab:p_ablation}
\end{table}

\myPara{Ablation Study on Fis-V.}
The ablation study conducted on the Fis-V dataset provides valuable insights into the individual contributions of core components within CoFInAl. The results in \cref{tab:ablation_fiv} showcase the performance impact of removing specific elements from the proposed framework.
Removing the regularization term leads to a notable performance drop of 3\%, emphasizing the importance of this term in effective prototype learning during training, especially in scenarios with limited labeled data. The absence of TFM results in a substantial performance decrease of 71\%. This emphasizes the crucial role of temporal fusion in achieving accurate assessments. The removal of GPM leads to a significant performance degradation of 67\%. GPM parses features into coarse and fine components, aligning with the hierarchical assessment approach. The drastic drop underscores the necessity of hierarchical processing for effective action assessment. The absence of FGS results in a slight performance decrease of 3\%. FGS contributes to fine-grained assessment within each grade, showcasing its importance in discerning variations within grades for nuanced AQA.
In conclusion, the ablation study underscores the significance of each core component in CoFInAl, highlighting their collective impact on achieving state-of-the-art AQA performance on the Fis-V dataset. The results validate the effectiveness of the proposed hierarchical alignment strategy.

\begin{table}[h]
    \centering
    \small
    \begin{tabular}{llllll}
    \toprule
    Setting & TES & PCS & Average \\
    \midrule
    Ours  & 0.716 & 0.843 & 0.788  \\
    \hline
    ~~w/o $\mathcal{L}_{\mathrm{R}}$ & 0.712$^{\downarrow 1\%}$ &  0.759$^{\downarrow 10\%}$  & 0.763$^{\downarrow 3\%}$  \\
    ~~w/o TFM & 0.150 $^{\downarrow 79\%}$ & 0.308 $^{\downarrow 63\%}$ & 0.231 $^{\downarrow 71\%}$   \\
    ~~w/o GPM & 0.301 $^{\downarrow 58\%}$ & 0.208 $^{\downarrow 75\%}$ & 0.260 $^{\downarrow 67\%}$   \\
    ~~w/o FGS  &  0.703$^{\downarrow 2\%}$  & 0.811$^{\downarrow 4\%}$ & 0.762$^{\downarrow 3\%}$   \\
    \bottomrule
    \end{tabular}
    \caption{Ablation results on the Fis-V dataset.}
    \label{tab:ablation_fiv}
\end{table}

\begin{figure}
    \centering
    \includegraphics[width=\linewidth]{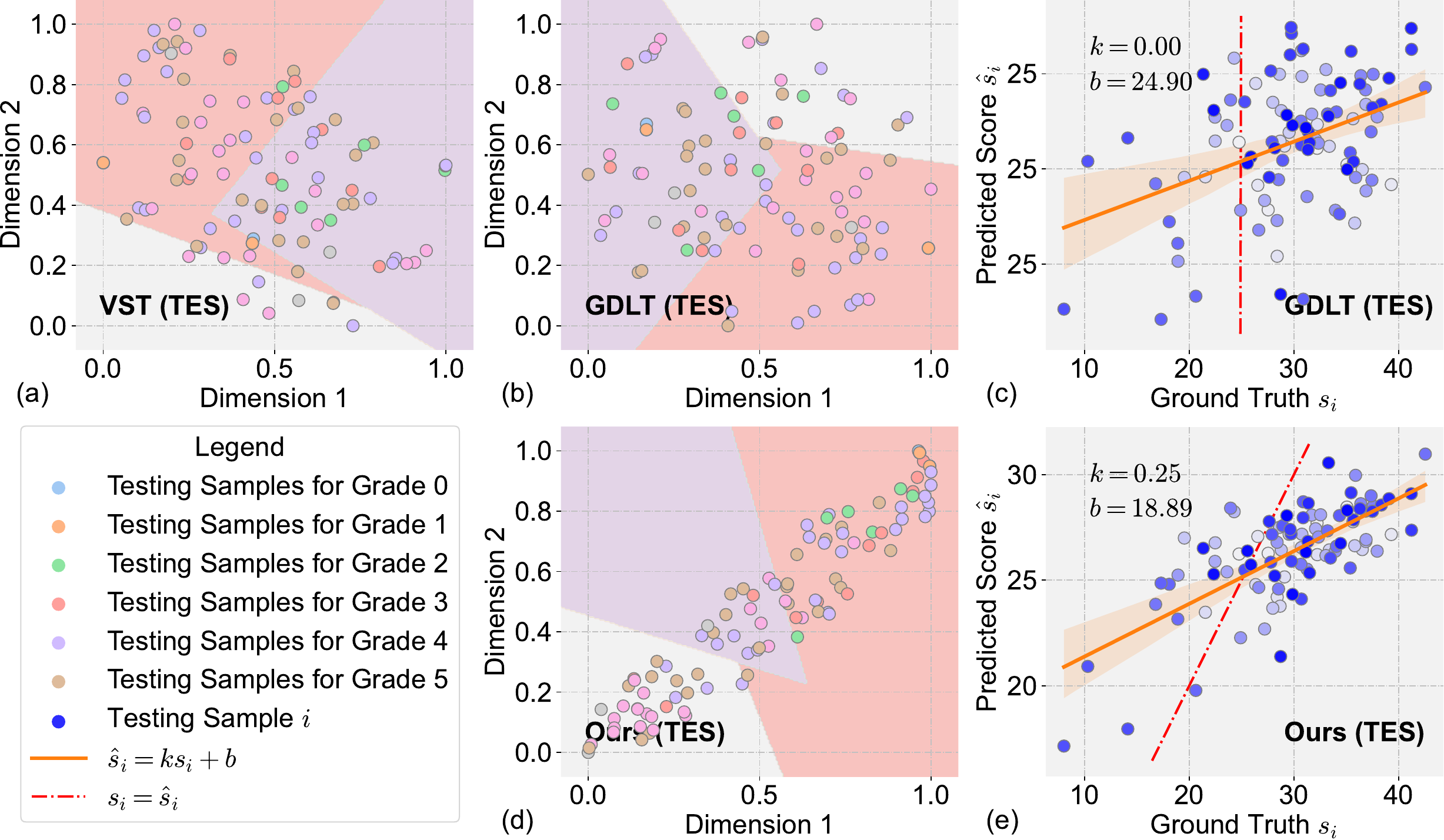}
    \caption{T-SNE feature distribution plots (a, b, d) and correlation comparison plots (c, e) contrasting GDLT with our CoFInAl method.}
    \label{fig:tsne_tes}
\end{figure}

\myPara{Effectiveness in Addressing Domain Shifts.}
We supplement another T-SNE feature distribution and correlation plots on TES (Fis-V), as shown in \cref{fig:tsne_tes}. The t-SNE feature distribution plot for TES testing samples categorized into grades reveals that the VST backbone produces scattered representations across grades, confirming the presence of domain discrepancies. In contrast, CoFInAl yields significantly more clustered embeddings demonstrating successful adaptation.
Further, the correlation plot comparing predicted and ground truth TES displays a higher slope for CoFInAl versus GDLT, aligning better with the ideal function. This implies enhanced score resolution from improved generalization.
Across both PCS and TES metrics associated with differing intricacies, CoFInAl consistently achieves effective domain translation, quantified by feature alignments and scoring accuracy improvements. The additional analysis underscores the wider applicability of precise athletic assessments involving complex temporal dynamics beyond holistic components.

\begin{figure}
    \centering
    \includegraphics[width=\linewidth]{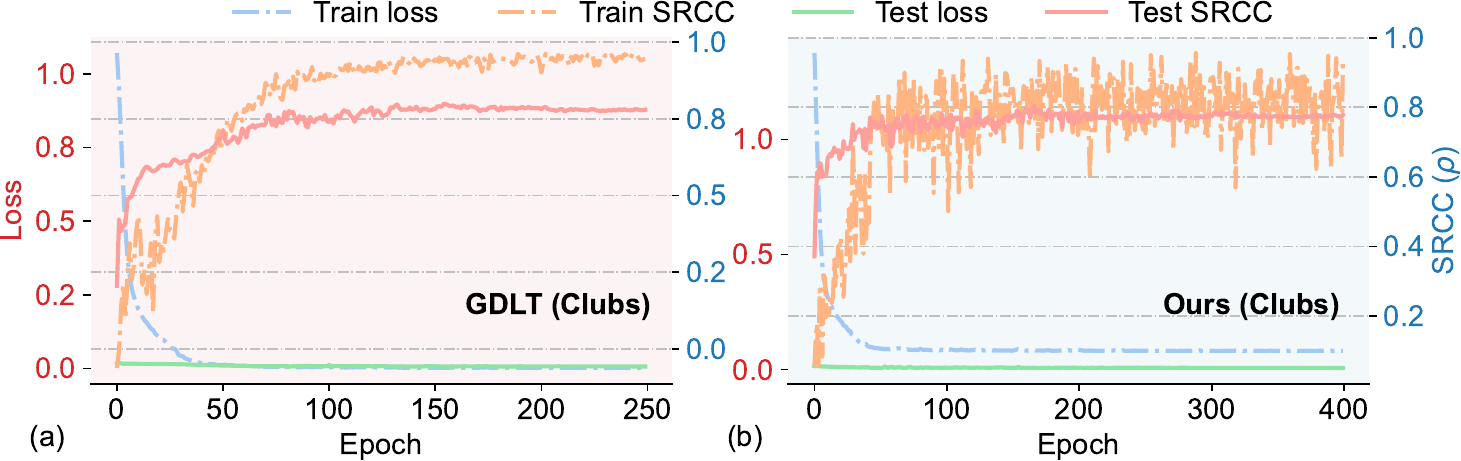}
    \caption{Comparison of loss and SRCC with GDLT on Clubs (RG).}
    \label{fig:loss_clubs}
\end{figure}
\begin{figure}
    \centering
    \includegraphics[width=\linewidth]{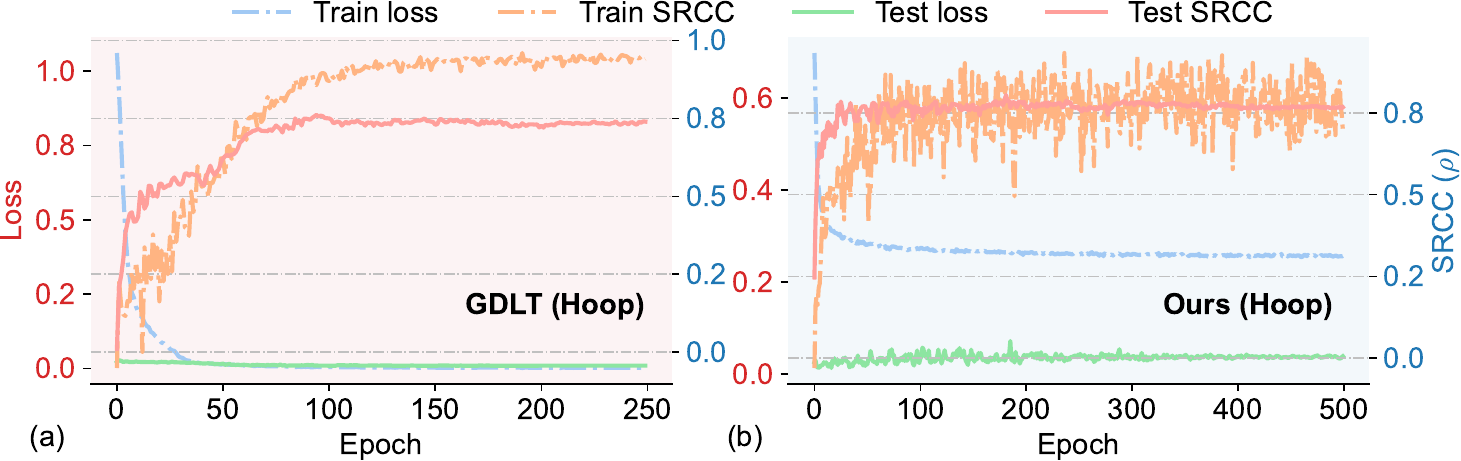}
    \caption{Comparison of loss and SRCC with GDLT on Hoop (RG).}
    \label{fig:loss_hoop}
\end{figure}
\begin{figure}
    \centering
    \includegraphics[width=\linewidth]{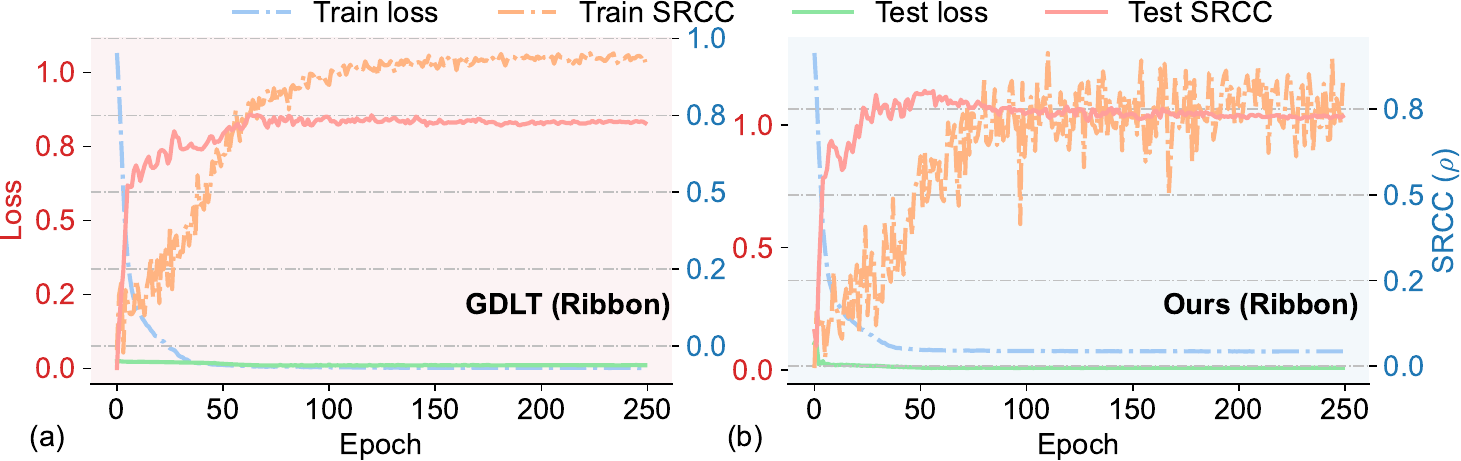}
    \caption{Comparison of loss and SRCC with GDLT on Ribbon (RG).}
    \label{fig:loss_ribbon}
\end{figure}

\myPara{Effectiveness of Mitigating Overfitting.} 
\cref{fig:loss_clubs,fig:loss_hoop,fig:loss_ribbon} illustrate the loss and SRCC comparison with GDLT \cite{xu2022likert} on Clubs, Hoop, and Ribbon (RG). 
The consistent outperformance of testing SRCC over training across categories emphasizes CoFInAl's effectiveness in aligning pre-training objectives with fine-tuning tasks. By avoiding overfitting to the training data, it facilitates reliable generalization to the testing set.
On the other hand, GDLT's discrepancy between training and testing performance highlights the inability of conventional fine-tuning to fully adapt models for the target task. Without effectively addressing domain shift issues, overfitting risks persist.
In summary, these additional ablation experiments further solidify CoFInAl's strengths in mitigating overfitting through principled alignment of instructions. 